\documentclass[a4paper,twoside]{article}

\usepackage[utf8]{inputenc}
\usepackage[T1]{fontenc}
\usepackage{epsfig}
\usepackage{subcaption}
\usepackage{calc}
\usepackage{amssymb}
\usepackage{amstext}
\usepackage{amsmath}
\usepackage{amsthm}
\usepackage{multicol}
\usepackage{pslatex}
\usepackage{apalike}
\usepackage{algorithm2e}
\usepackage[bottom]{footmisc}
\usepackage{bm}
\usepackage{graphicx}
\usepackage{booktabs}
\usepackage{microtype}
\usepackage[hidelinks]{hyperref}
\usepackage[sort]{cleveref}
\usepackage{SCITEPRESS}

\crefname{equation}{Equation}{Equations}
\crefname{figure}{Figure}{Figures}
\crefname{table}{Table}{Tables}
\crefname{section}{Section}{Sections}
\crefname{subfigure}{Figure}{Figures}

\hypersetup{pdftitle={Beyond the Vocabulary: Interpreting Visual Tokens in Large-Scale Vision-Language Models},pdfauthor={},pdfsubject={},pdfkeywords={}}

\begin{document}

\title{Beyond the Vocabulary: Interpreting Visual Tokens \\ in Large-Scale Vision-Language Models}

\author{
\authorname{
Yuji Kitano\sup{*},
Yuto Nishida\sup{*},
Haruki Sakajo,
Hidetaka Kamigaito
and Taro Watanabe
}
\affiliation{Division of Information Science, Nara Institute of Science and Technology, Nara, Japan}
\email{\{kitano.yuji.la6, nishida.yuto.nu8, sakajo.haruki.sd9, kamigaito.h\}@naist.ac.jp, taro@is.naist.jp}
}


\abstract{
Large-scale vision-language models (LVLMs) process visual and textual tokens through the shared Transformer layers that generate language, yet it remains unclear what visual information is represented inside the language model (LM) component and how these representations interact across layers.
Prior work often interprets visual tokens through language-centric tools such as the logit lens, which decode visual representations as text tokens, whereas non-linguistic structure, such as token position, can be difficult to characterize in vocabulary space.
We introduce an Independent Component Analysis (ICA)-based framework for analyzing visual token representations across LM layers without projecting them into vocabulary space.
Our analyses across different LVLM architectures reveal ICA axes associated with object categories, token positions, and, qualitatively, color, with systematic, model-dependent layerwise variation in the prevalence of object- and position-selective axes.
These results show that the organization of visual-token representations along interpretable semantic and spatial axes changes across LM layers, highlighting ICA as a complementary lens for analyzing visual tokens beyond vocabulary-space projections.
}

\onecolumn
\maketitle

\begingroup
\renewcommand{\thefootnote}{\fnsymbol{footnote}}
\footnotetext[1]{These authors contributed equally to this work.}
\endgroup

\normalsize
\setcounter{footnote}{0}
\vfill

\section{INTRODUCTION}
\label{sec:introduction}

Large-scale vision-language models (LVLMs) have become a standard paradigm for building multimodal systems that can describe images, answer visual questions, read text in images, and reason over visual inputs~\cite{liu2023visual,bai2025qwen3vltechnicalreport,gemmateam2025gemma3technicalreport}.
A typical design in these models combines a language model (LM) with a vision encoder and integrates the resulting visual features into the LM as visual tokens.
These visual tokens are then processed through the LM's Transformer stack, which ultimately supports language generation.
This design has enabled strong performance, but it also raises a basic interpretability question: what visual information is represented inside the LM residual stream, and how are these representations transformed across layers? 

Existing work has made substantial progress in understanding how LVLMs use visual information~\cite{neo2025towards,kaduri-et-al-2025-deep-dive,venhoff2025visualrepresentationsmaplanguage,zhang2025crossmodal,serra2025the}.
To interpret visual representations in residual streams, prior work~\cite{neo2025towards} employs the logit lens~\cite{logitlens} and shows that those representations are interpretable through LMs' vocabulary in the middle-to-late layers.
However, visual factors such as token position, spatial layout, and local appearance may not naturally correspond to individual vocabulary tokens.
Prior work~\cite{phukan-etal-2025-beyond} has also identified limitations of logit-lens-based probing for contextual visual relationships, including attributes and spatial relations.
This motivates a complementary approach that analyzes visual-token representations without restricting interpretation to vocabulary-defined directions.

We introduce an Independent Component Analysis (ICA)-based framework for interpreting visual tokens in LVLMs beyond the vocabulary space. 
ICA seeks statistically independent components through linear transformations of high-dimensional representations~\cite{HYVARINEN2000411}, and has been shown to reveal interpretable semantic axes in word and image embeddings~\cite{yamagiwa-etal-2023-discovering}.
Given an LM, we collect hidden states at visual token positions and apply ICA to the resulting residual stream representations for each LM layer. 
We interpret the resulting components using visual-factor metadata, including object-category annotations, token positions, and local color information.
This modality-agnostic framework does not require vocabulary-based decoding.
We apply this framework to LVLMs with different architectures using images from the COCO validation set.
We present each image without a text prompt and compare visual token representations across LM layers to examine how the organization of visual information differs across models.

Our analysis reveals that the LM residual stream contains linearly separable visual factors that are not limited to vocabulary-level semantics. 
We identify components associated with object categories and token positions across three LVLMs, as well as qualitative examples of color-related components.
The fractions of factor-selective ICA axes exhibit systematic, model-dependent layerwise profiles. Object-selective axes become less prevalent from shallow to intermediate layers, with different trends in deeper layers across models, while position-selective axes increase markedly in deeper Qwen layers.
These findings demonstrate that ICA provides a complementary view of visual-token representations, revealing model-dependent layerwise variation in the prevalence of semantic- and spatial-selective axes.

\section{BACKGROUND}
\label{sec:background}

\subsection{Residual Streams}

In Transformer~\cite{vaswani2017attention} with the pre-layer normalization architecture~\cite{xiong2020onlayer}, hidden states gradually evolve layer by layer, forming \emph{residual streams}~\cite{mathematicalframework}.

Let $\bm{H}^{(\ell)} \in \mathbb{R}^{n \times d}$ denote the hidden states of $n$ tokens at layer $\ell$, where $d$ is the hidden dimension. 
At $(\ell +1)$-th layer, the update of the hidden state through $(\ell + 1)$-th Transformer block $f^{(\ell + 1)}_\theta: \mathbb{R}^{n \times d} \rightarrow \mathbb{R}^{n \times d}$ can be written as:
\begin{equation}
\label{eq:residual-stream}
    \bm{H}^{(\ell + 1)} = \bm{H}^{(\ell)} + f^{(\ell + 1)}_\theta(\bm{H}^{(\ell)})
\end{equation}

Thus, each layer updates the existing token representations rather than replacing them entirely. 
This residual stream view provides a natural object of analysis for studying how token representations are transformed across layers.

\subsection{Large-Scale Vision-Language Models}
Large-scale vision-language models (LVLMs)~\cite{liu2023visual,bai2025qwen3vltechnicalreport,gemmateam2025gemma3technicalreport} extend language models (LMs) to condition on both visual and text inputs.
A common architecture encodes an image with a vision encoder, maps the resulting visual features into the LM's embedding space, and inserts them into the LM input sequence as visual tokens. 
Once visual tokens enter the LM stack, they are processed by the same Transformer layers that support language generation. 

\subsection{Interpretability for Language Models}

Interpretability research on language models examines activations in residual streams and Transformer submodules such as multi-head attention and feed-forward networks.
Sparse autoencoders (SAEs), transcoders, and circuit analyses based on them help explain how LMs process representations~\cite{huben2024sparse,dunefsky2024transcoders}.
The Logit Lens and Independent Component Analysis (ICA) characterize the static, compositional features of residual stream representations~\cite{yamagiwa-etal-2023-discovering,oyama-etal-2024-understanding,wendler-etal-2024-llamas}.
Although the Logit Lens successfully projects interpretable, symbol-level textual data, this surface-level mapping inherently precludes the extraction of low-level multimodal features, such as spatial positioning or chromatics.

Prior work on LVLMs uses the Logit Lens, SAEs, and attention knockout to probe cross-modal information flow during text generation~\cite{liu2025mechanistic,neo2025towards,kaduri-et-al-2025-deep-dive,zhang2025crossmodal,serra2025the} and interpret visual representations using the LM's vocabulary~\cite{neo2025towards} or features~\cite{venhoff2025visualrepresentationsmaplanguage}.
These methods help identify when visual information becomes accessible for generation, but often emphasize vocabulary-aligned or generation-facing features~\cite{phukan-etal-2025-beyond}.
Which visual properties the tokens encode remains less clear. We therefore examine vocabulary-aligned semantic factors and non-linguistic visual factors together in the LM residual stream.

\section{ICA-BASED ANALYSIS OF VISUAL TOKENS}
\label{sec:method}

For each model, we fit ICA separately to the residual stream activations at visual token positions in each LM layer.
We interpret each axis by mapping its highest-scoring tokens to the image regions they represent and examining the visual properties those regions share.

\subsection{Visual Tokens in the LM Residual Stream}
\label{sec:visual_tokens}

Most LVLMs convert an input image into a sequence of visual tokens and insert those tokens into the LM input sequence together with text tokens. 
Visual and text tokens pass through the same Transformer layers in the LM.
We therefore analyze the residual stream at visual token positions after each layer (\cref{eq:residual-stream}).

Let $\{I_1, \ldots, I_M\}$ denote a set of input images. 
For image $I_m$, let $N_m$ be the number of visual tokens produced by the model-specific preprocessing and visual tokenization pipeline. 
We denote the hidden state of the $n$-th visual token in image $I_m$ at LM layer $\ell$ as
\begin{equation}
    \bm{h}^{(\ell)}_{m,n} \in \mathbb{R}^{d},
    \qquad
    m = 1,\ldots,M,\quad n = 1,\ldots,N_m,
\end{equation}
where $d$ is the hidden dimensionality of the LM. 
For each layer $\ell$, we concatenate all visual token hidden states across images into a single activation matrix:
\begin{equation}
    \bm{H}^{(\ell)}
    =
    \begin{bmatrix}
    \bm{h}^{(\ell)}_{1,1} \\
    \vdots \\
    \bm{h}^{(\ell)}_{M,N_M}
    \end{bmatrix}
    \in \mathbb{R}^{N \times d},
    \qquad
    N = \sum_{m=1}^{M} N_m .
\end{equation}
Each row of $\bm{H}^{(\ell)}$ corresponds to one visual token representation, and each column corresponds to one hidden dimension. 
We use this layer-specific population of visual token representations to analyze how semantic, spatial, and appearance-related factors are organized in the LM residual stream across layers.

\subsection{ICA-Based Interpretation of Visual Token Activations}
\label{sec:ica_interpretation}

Given the layer-specific activation matrix $\bm{H}^{(\ell)}$, we apply independent component analysis (ICA) separately to each LM layer. 
ICA produces $k$ independent axes over the visual token population and assigns a component score to every visual token. 
We denote the resulting score matrix by $\bm{S}^{(\ell)} \in \mathbb{R}^{N \times k}$, where $S^{(\ell)}_{t,i}$ is the score of visual token $t$ on component $i$ at layer $\ell$.

We interpret each component through the visual tokens that most strongly activate it. 
Following prior analyses of representation-space directions, where highly activating examples often provide the clearest basis for interpretation, we use the top-scoring tokens as evidence for what visual factor a component captures. 
For component $i$ at layer $\ell$, let
\begin{equation}
    \mathcal{T}^{(\ell,i)}_K
    =
    \operatorname{TopK}
    \left(
    \bm{S}^{(\ell)}_{:,i}
    \right)
\end{equation}
denote the set of top-$K$ visual tokens for that component. 
Because the sign of ICA components is arbitrary, we examine both positive and negative directions.

We map these top-activating tokens back to their corresponding image regions and characterize them using token-level visual factors. 
Object labels, token coordinates, and color descriptors are used only after ICA fitting to interpret the components.
This protocol provides a common way to examine whether ICA axes are associated with vocabulary-aligned factors or with non-linguistic visual structure.

\begin{figure*}[t]
    \centering

    \begin{subfigure}[t]{\textwidth}
        \centering
        \includegraphics[width=0.48\textwidth]{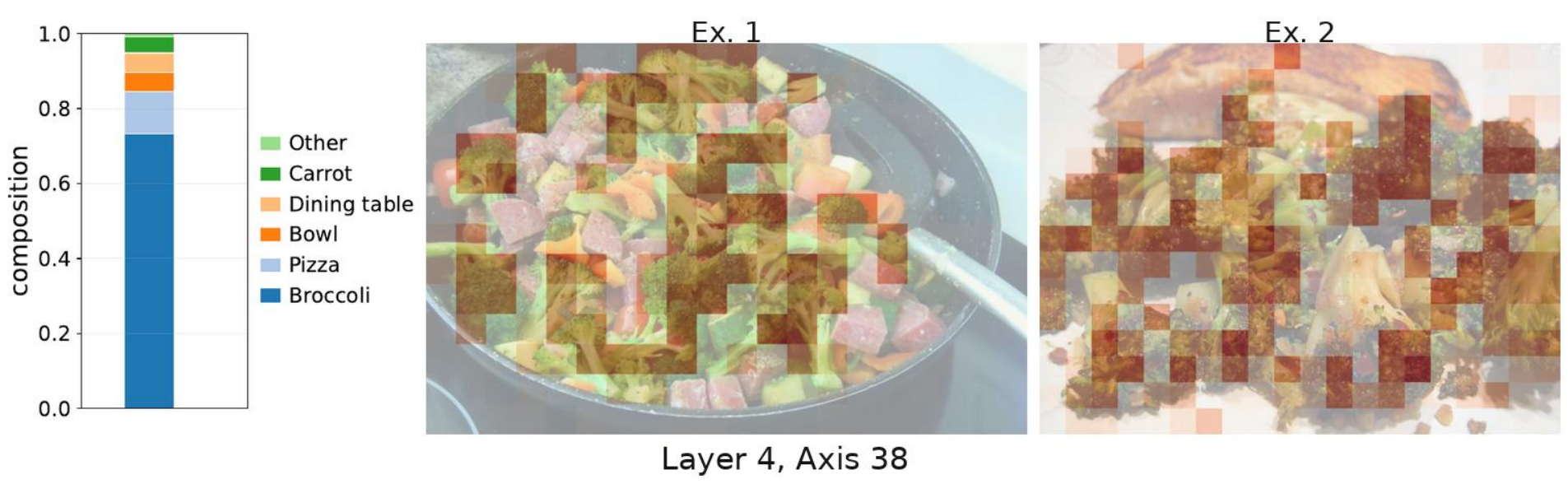}
        \hfill
        \includegraphics[width=0.48\textwidth]{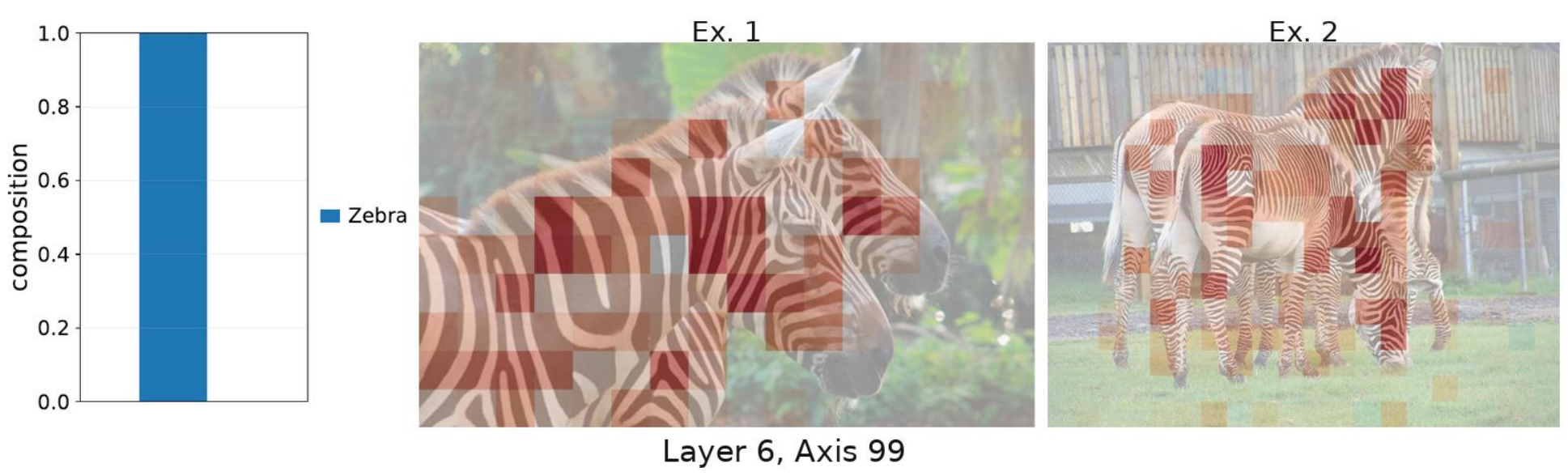}
        \caption{
Object-related axes.
The summary plot shows the object-category composition of the top-activating tokens; the examples show representative high-scoring image regions.
Here, the left axis is selective for Broccoli, whereas the right axis is selective for Zebra.
}
        \label{fig:ica_examples_object}
    \end{subfigure}

    \vspace{0.5em}

    \begin{subfigure}[t]{\textwidth}
        \centering
        \includegraphics[width=0.48\textwidth]{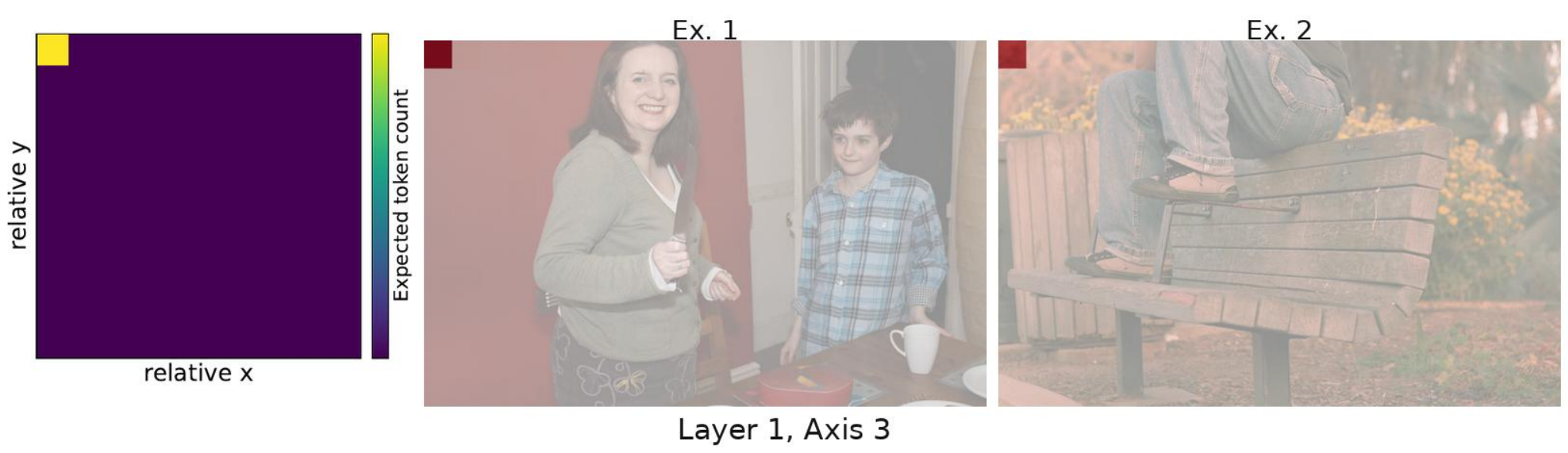}
        \hfill
        \includegraphics[width=0.48\textwidth]{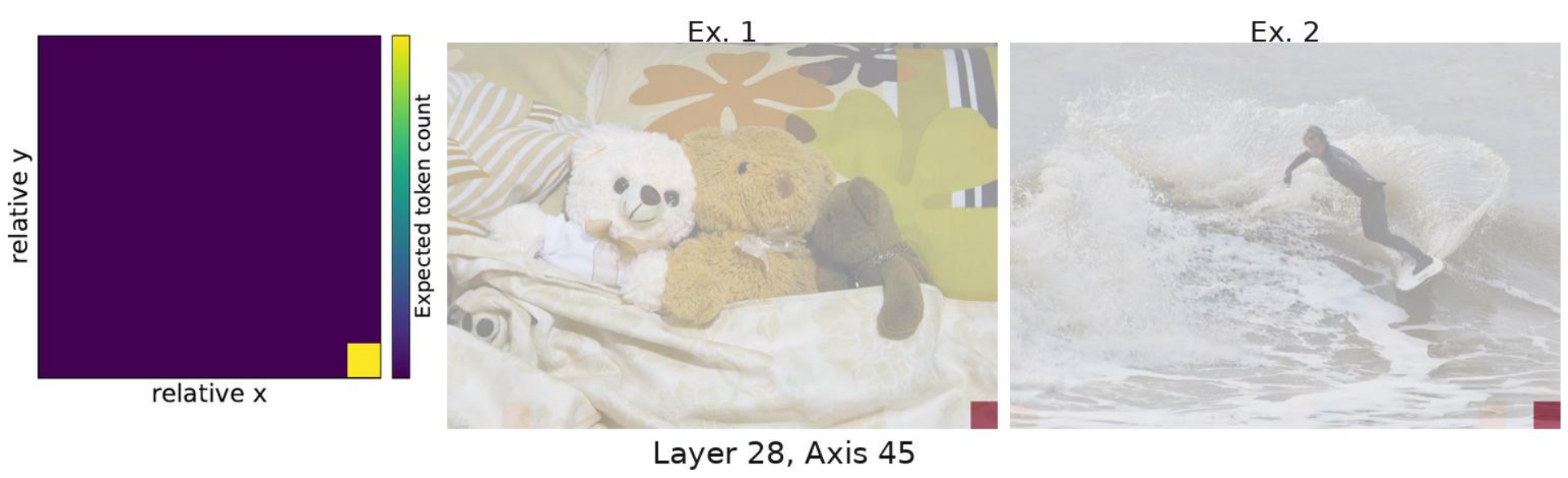}
\caption{
Position-related axes.
The summary heatmap aggregates the relative coordinates of top-activating tokens; warmer regions indicate where high-scoring tokens concentrate.
Here, the left axis responds to tokens near the upper left, whereas the right axis responds to tokens near the lower right.
}
        \label{fig:ica_examples_position}
    \end{subfigure}

    \vspace{0.5em}

    \begin{subfigure}[t]{\textwidth}
        \centering
        \includegraphics[width=0.48\textwidth]{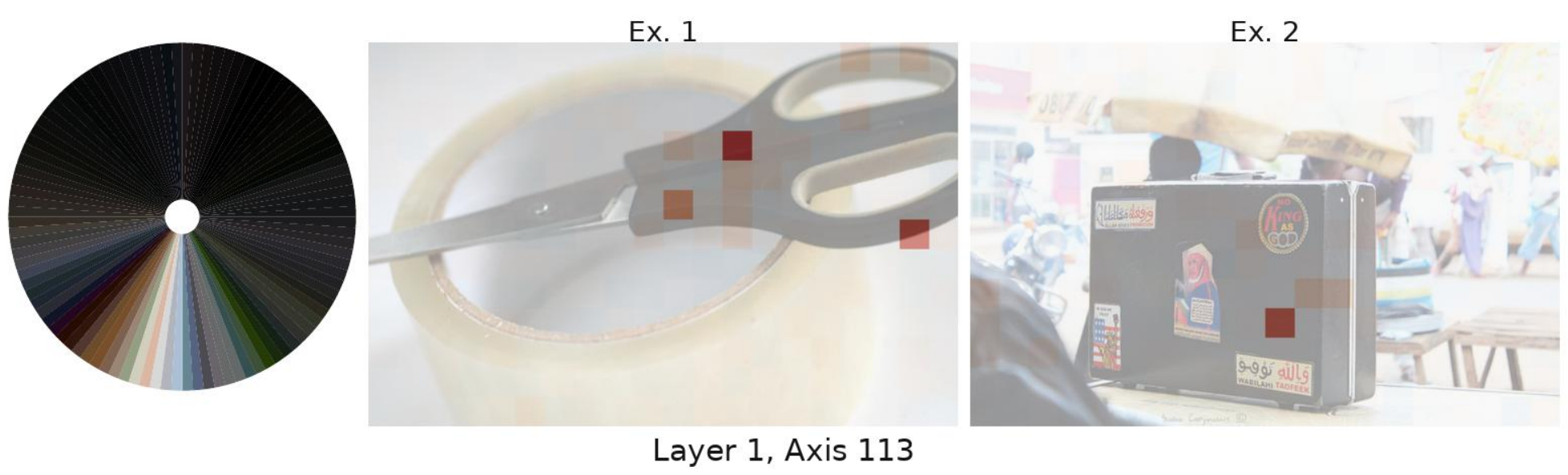}
        \hfill
        \includegraphics[width=0.48\textwidth]{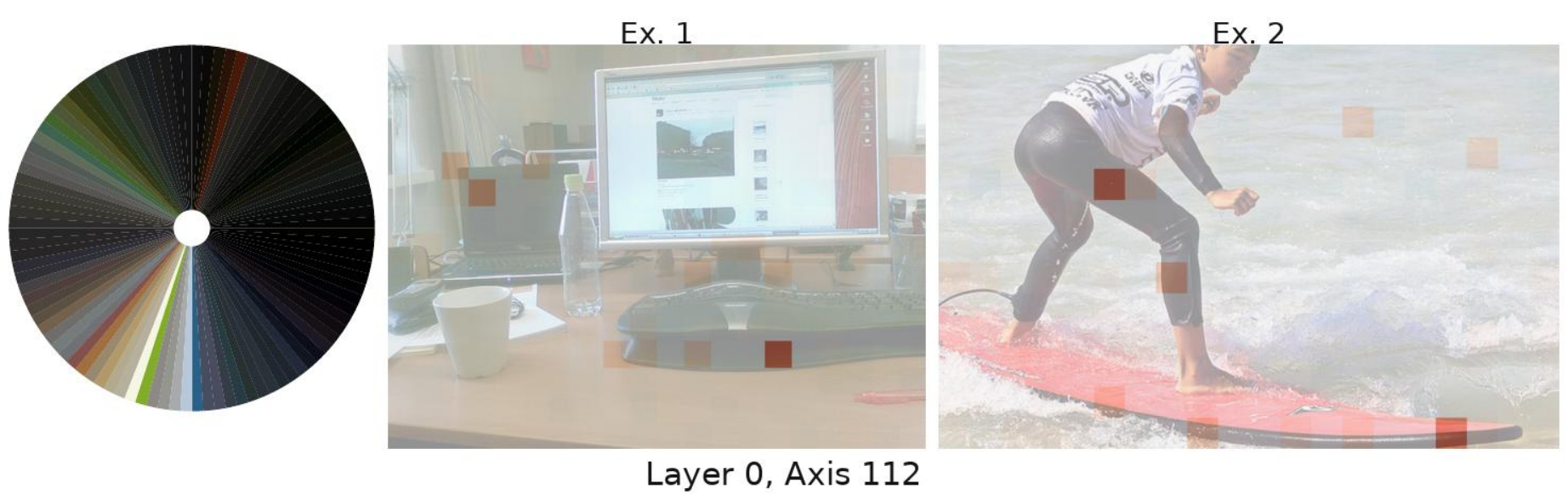}
\caption{
Color-related axes.
The summary ring shows the colors of top-activating token regions, ordered by color similarity; the examples show high-scoring regions with similar appearance.
Here, both axes are selective for black regions.
}
        \label{fig:ica_examples_color}
    \end{subfigure}

    \captionsetup{skip=10pt}
    \caption{
    Representative ICA axes in the LM residual stream of the primary model, Qwen3-VL-8B-Instruct.
    Each row shows two axes associated with the same type of visual factor.
    For each axis, the left panel summarizes its top-activating visual tokens, and the two panels on the right show representative top-activating examples with token-level activation heatmaps.
    The examples illustrate that ICA exposes interpretable semantic, spatial, and appearance-related axes in a common representation space.
    }
    \label{fig:ica_examples}
\end{figure*}

\section{EXPERIMENTS}
\label{sec:experiments}

We examine whether ICA reveals both vocabulary-aligned semantic factors, such as object category, and non-linguistic visual factors, such as token position and color, in a common component space.
We then ask how the prevalence of object- and position-selective axes varies with layer depth and whether these patterns differ across LVLMs.

\subsection{Experimental Setup}
\label{sec:experimental_setup}

\paragraph{Models.}
Our primary model, Qwen3-VL-8B-Instruct~\cite{bai2025qwen3vltechnicalreport}, uses dynamic resolution and merges $2\times2$ neighboring visual patches. We record the resulting token count for each image.
LLaVA-1.5-7B~\cite{liu2024improved} encodes $336\times336$ images with CLIP ViT-L/14 and projects the $24\times24$ feature grid into the LM with a two-layer MLP.
Only the projector is trained during LLaVA's feature-alignment stage.
Gemma-3-4B-it~\cite{gemmateam2025gemma3technicalreport} uses bidirectional attention among tokens from the same image. We disable Pan\&Scan cropping and use a single $896\times896$ image per input.
Table~\ref{tab:model_dimensions} summarizes the resulting token counts and LM dimensions. We use the released checkpoints and HuggingFace Transformers implementations~\cite{wolf-etal-2020-transformers}.

\begin{table}[t]
\centering
\caption{LM dimensions and visual tokens per image. Qwen's count is the mean over the 740 sampled images.}
\label{tab:model_dimensions}
\small
\begin{tabular*}{\columnwidth}{@{\extracolsep{\fill}}lrrr@{}}
\toprule
Model & LM layers & Hidden size & Tokens/image \\
\midrule
Qwen  & 36 & 4096 & 265 (mean) \\
LLaVA & 32 & 4096 & 576 \\
Gemma & 34 & 2560 & 256 \\
\bottomrule
\end{tabular*}
\end{table}

\paragraph{Dataset and annotations.}
We use the COCO 2017 validation dataset~\cite{mscoco}, which contains natural color images with object-category annotations, segmentation masks for annotated object regions, and human-written captions. 
Captions are used only to construct the evaluation subset and are never provided to the models. 
We restrict our analysis to the 74 COCO object categories whose category names are explicitly mentioned in at least 10 validation captions. 
For each analyzed category, we randomly sample 10 distinct color images with a fixed random seed, resulting in 740 images in total.

For each model, we associate every visual token with the image region from which it is derived under the model-specific visual tokenization pipeline. 
Using this token-to-region correspondence, we assign object and spatial information to each visual token. 
Object labels are derived from the COCO segmentation annotations. 
For each visual token, we compute the proportion of its corresponding image region that overlaps with each of the 74 analyzed COCO object categories; the token is assigned the category with the largest overlap, or background if it does not overlap any analyzed category, yielding 75 possible token labels in total. 
We also record the spatial coordinates of each token region in the processed image, which allows us to analyze position-related components in the LM residual stream. 
When analyzing appearance-related factors, we additionally compute simple descriptors from the corresponding image region, such as average color.

\paragraph{Representation extraction.}
Each input consists only of a single COCO image, with no textual instruction, object-category name, or COCO caption provided to the model.\footnote{In the standard input formats of the evaluated models, the user instruction follows the image tokens. Under this image-first ordering, post-image text does not affect the preceding visual-token states analyzed here.}
Each image is processed using the model's standard vision tokenization pipeline. 
For each LM layer, we extract the residual stream hidden states at visual token positions. 
When a model inserts special image boundary or placeholder tokens, such as image-start or image-end tokens, we exclude them and retain only the hidden states corresponding to actual visual tokens. 
The resulting layerwise visual token representations are used to construct the activation matrices for ICA.

\paragraph{ICA implementation details.}
We apply FastICA~\cite{HYVARINEN2000411} separately for each model and each LM layer. 
The main analysis uses $k=740$ components, matching the number of sampled images.
We also consider $k=74$, the number of object categories, and $k$ equal to each model's mean visual token count per image (Table~\ref{tab:model_dimensions}).
We center and whiten each activation matrix and run FastICA for up to 10,000 iterations with a tolerance of $10^{-10}$, following prior work~\cite{yamagiwa-etal-2023-discovering}.
For layers that reach this limit without converging, we use the solution from the iteration with the highest similarity between the transformation matrices immediately before and after the update.
Unless otherwise stated, component-level analyses use the top-100 visual tokens with the largest activation scores for each ICA component.

\begin{figure*}[t]
\centering

\begin{subfigure}[t]{0.50\linewidth}
    \centering
    \includegraphics[width=\linewidth]{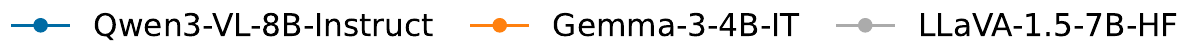}
\end{subfigure}
\\
\begin{subfigure}[t]{0.38\linewidth}
    \centering
    \includegraphics[width=\linewidth]{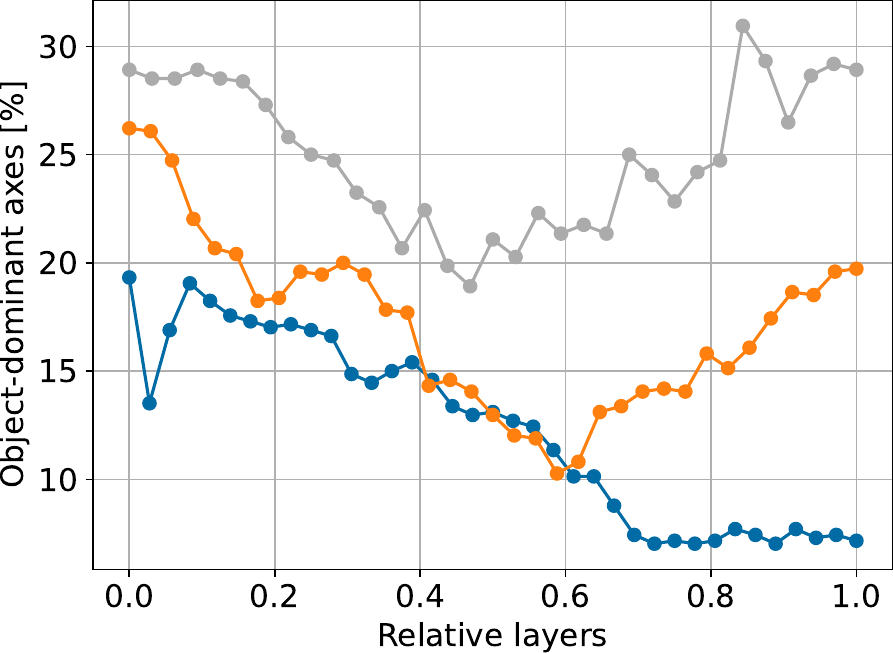}
    \caption{Object Category.}
    \label{fig:object_dominance_three_models}
\end{subfigure}
\hspace{3em}
\begin{subfigure}[t]{0.38\linewidth}
    \centering
    \includegraphics[width=\linewidth]{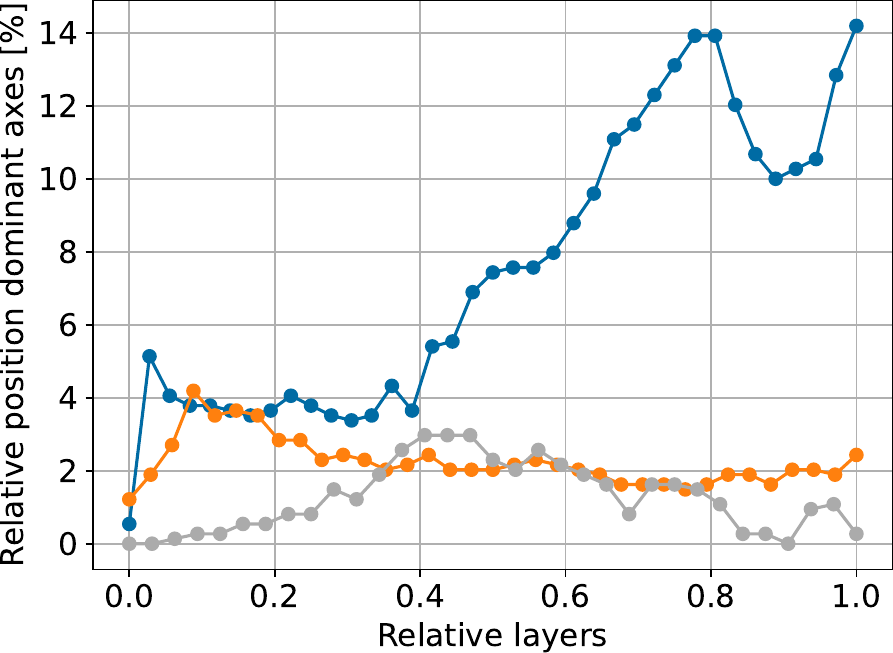}
    \caption{Token Position.}
    \label{fig:position_dominance_three_models}
\end{subfigure}
\vspace{0.2em}

\caption{Fraction of ICA axes selective for (a) object category and (b) token position across layers of Qwen, Gemma, and LLaVA ($k=740$).
An axis is selective if a majority of its top-100 activating tokens share an object category (a) or if those tokens cluster in a restricted image region (b).}
\label{fig:object_position_dominance_three_models}
\end{figure*}

\subsection{ICA Reveals Interpretable Visual Axes in the LM Component}
\label{sec:interpretable_axes}

We first examine whether ICA yields interpretable visual axes in the LM residual stream. 
For each component, we inspect the visual tokens with the largest activation scores and map them back to their corresponding image regions. 
Following prior representation-space analyses that interpret directions through highly activating examples, we use these top-activating visual tokens as evidence for the visual factor associated with each component. 
Because the sign of ICA components is arbitrary, we inspect both directions and show the direction with the clearer visual interpretation.

Figure~\ref{fig:ica_examples} shows representative ICA components from the primary model, with $k=740$. 
The examples indicate that ICA identifies axes associated with multiple types of visual factors. 
Some components are object-related: their top-activating tokens consistently correspond to the same object category or object part across different images. 
Other components are position-related: their activations concentrate in similar image regions across examples, showing that spatial position remains linearly separable in the LM residual stream. 
We also find color-related components, whose top-activating tokens share a similar appearance even when they belong to different object categories.

These examples show that ICA exposes not only vocabulary-aligned semantic directions. 
It also reveals non-linguistic visual structure, such as position and appearance, within the same component space. 
This motivates the following analyses, where we quantify how object- and position-related axes change across LM layers.

\subsection{Layerwise Selectivity Across Models}
\label{sec:axes_dynamics}

We next analyze how different types of visual factors evolve across LM layers and how their layerwise dynamics differ across LVLM architectures. 
We focus on two factors that instantiate the contrast introduced in our framework: object category as a vocabulary-aligned semantic factor and token position as a non-linguistic visual factor. 
For each model and each layer, we identify factor-selective ICA axes using the top-100 activating visual tokens of each component. 
An axis is counted as object-selective when a majority of its top tokens share the same object category, and as position-selective when the standard deviation of the relative coordinates of its top-100 activating visual tokens is smaller than the model-specific characteristic size of one visual-token region.

Figure~\ref{fig:object_position_dominance_three_models} summarizes the fraction of object- and position-selective axes across layers for Qwen, Gemma, and LLaVA.

\paragraph{Object categories.}
Figure~\ref{fig:object_dominance_three_models} shows that object-selective axes exhibit non-trivial layerwise dynamics rather than being uniformly preserved across the LM stack. 
Across models, object-category information is linearly separable in the residual stream, indicating that visual tokens retain vocabulary-aligned semantic structure after entering the LM layers. 
While object-selective axes are prominent in shallow layers for these three models, the behavior in middle and deeper layers differs across models.
Qwen exhibits lower fractions in deeper layers, and other models show high fractions, suggesting that each model encodes different information in residual streams.

\paragraph{Token positions.}
Unlike object categories, token position is a non-linguistic visual factor that is not naturally characterized by decoding a single vocabulary item, and \cref{fig:position_dominance_three_models} shows a different pattern for position-selective axes from object-dominant ones.
While Qwen gradually increases the proportion of position-dominant axes and exhibits substantially higher fractions in deeper layers, Gemma and LLaVA show weaker and more stable position-dominant profiles. This suggests that spatial-layout information remains especially salient in Qwen's LM residual stream, where it becomes concentrated into individual ICA axes. In contrast, the other models may retain positional information in a more distributed form that is less captured by our axis-level dominance criterion.

\begin{figure*}[t]
\centering
\includegraphics[width=0.45\linewidth]{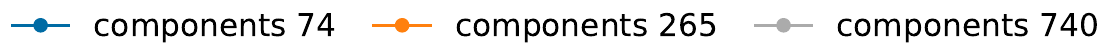}\\
\begin{subfigure}[t]{0.35\linewidth}
    \centering
    \includegraphics[width=\linewidth]{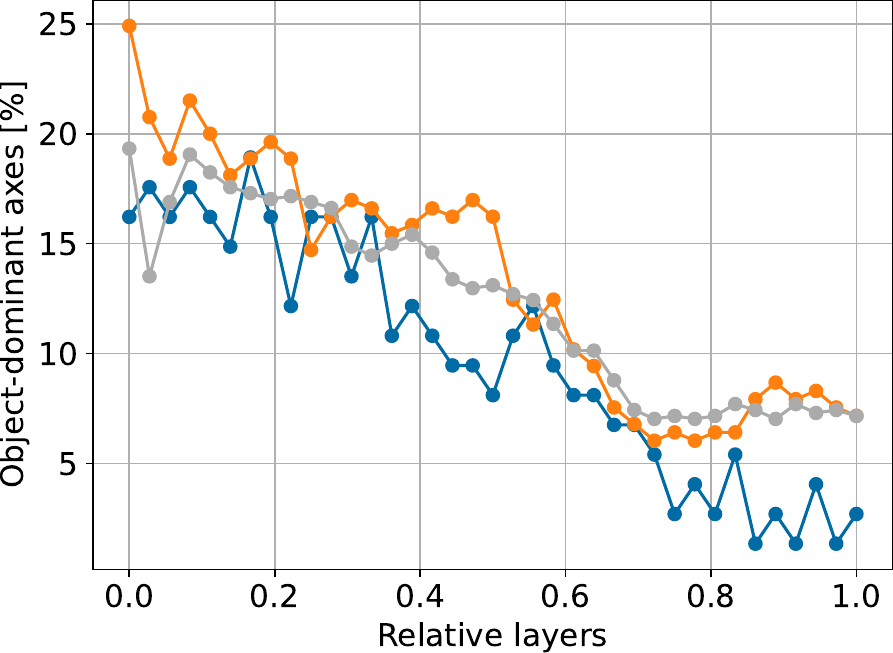}
    \caption{Object Category.}
    \label{fig:qwen_object_selective_n_components}
\end{subfigure}
\hspace{3em}
\begin{subfigure}[t]{0.35\linewidth}
    \centering
    \includegraphics[width=\linewidth]{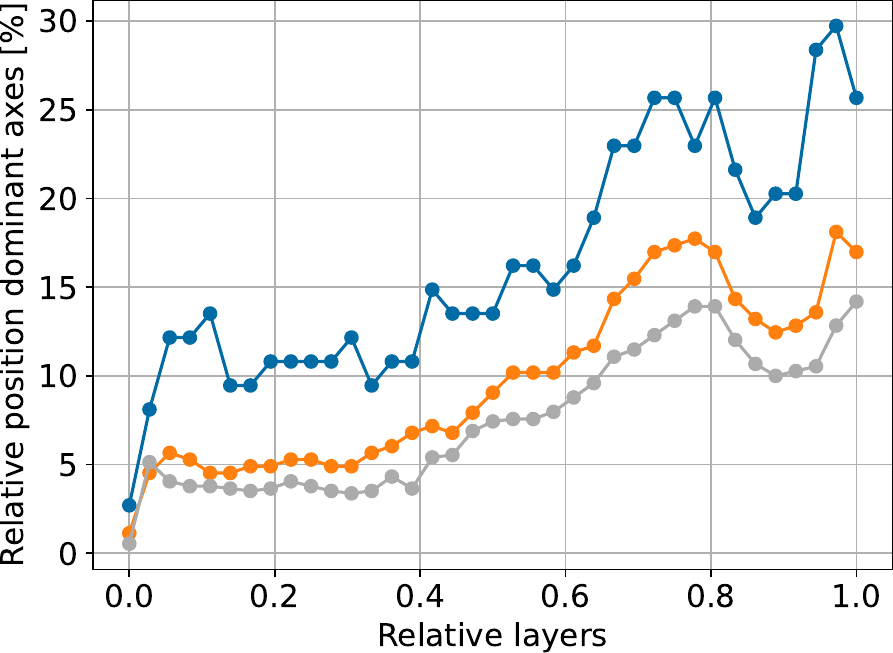}
    \caption{Token Position.}
    \label{fig:qwen_position_selective_n_components}
\end{subfigure}\\[0.6em]
\includegraphics[width=0.5\linewidth]{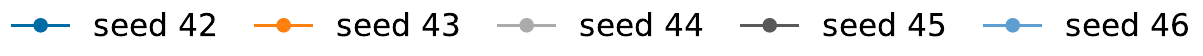}\\
\begin{subfigure}[t]{0.35\linewidth}
    \centering
    \includegraphics[width=\linewidth]{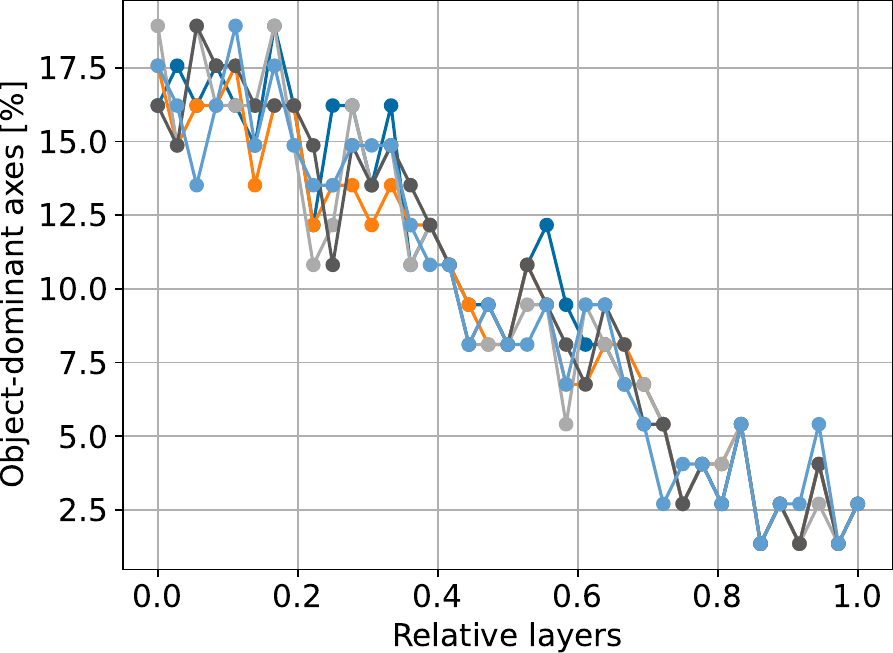}
    \caption{Object Category.}
    \label{fig:qwen_object_selective_seed_k74}
\end{subfigure}
\hspace{3em}
\begin{subfigure}[t]{0.35\linewidth}
    \centering
    \includegraphics[width=\linewidth]{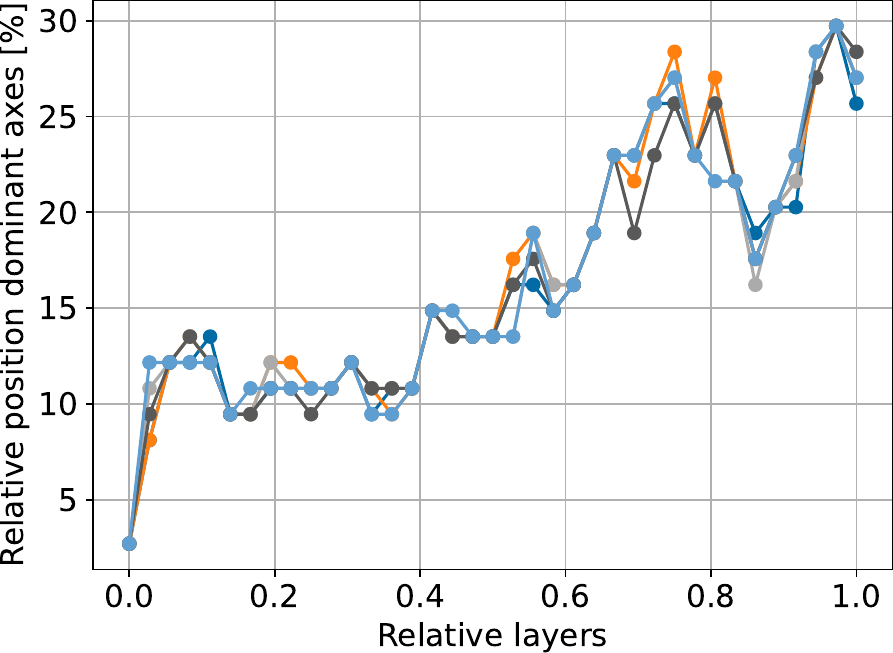}
    \caption{Token Position.}
    \label{fig:qwen_position_selective_seed_k74}
\end{subfigure}
\vspace{0.2em}
\caption{Qwen layerwise selectivity (\%). With seed 42 (top), the number of images, the number of object categories, and the average number of visual tokens per image give $k=740,74,265$, respectively. Seeds vary at $k=74$ (bottom).}
\label{fig:qwen_ica_ablation}
\end{figure*}

\subsection{Validation and Ablation Studies}
\label{sec:validation_ablation}

We validate that the main layerwise trends are not artifacts of a particular ICA configuration.
Specifically, we examine the robustness of the Qwen results to two sources of variation: the number of independent components and the random seed used for ICA initialization.
We apply the same selectivity criteria to the top 100 activating tokens per axis (Section~\ref{sec:axes_dynamics}).

The absolute fractions vary with decomposition granularity (Figure~\ref{fig:qwen_object_selective_n_components}--\subref{fig:qwen_position_selective_n_components}), and both profiles are similar across seeds 42--46 at $k=74$ (Figure~\ref{fig:qwen_object_selective_seed_k74}--\subref{fig:qwen_position_selective_seed_k74}). Individual components may change across fits, yet object-selective axes remain more frequent in shallow layers and position-selective axes in deeper layers.

\section{DISCUSSION}
\label{sec:discussion}

\subsection{Visual Tokens Across LM Layers}
\label{sec:discussion_LM_image_tokens}

A common view of LVLMs is that the vision encoder extracts visual features, the projector maps them into the LM embedding space, and the LM generates text conditioned on those features.
This description leaves open how the visual representations change before generation, as visual tokens pass through the same Transformer stack as text tokens.
The layerwise ICA profiles provide a way to examine this intermediate organization.

The fraction of object-selective axes decreases from shallow to intermediate layers, then increases again in deeper layers of Gemma and LLaVA while remaining low in Qwen.
One possible interpretation is that object information becomes more distributed in intermediate layers and hence less apparent along individual ICA directions. In Gemma and LLaVA, object-specific structure may become more explicit again in layers closer to the language output.
The fraction of selective axes describes how ICA captures object information at each layer. A decrease alone cannot establish that object information has been lost.

The position-selective axes show that the LM residual stream also retains information about image layout. In Qwen, such axes become more prevalent in deeper layers, where relatively few axes are object-selective.
The qualitative color examples show that low-level appearance can also remain linearly separable alongside object category and position.
Since a visual property shared by several image regions need not correspond to a single likely output word, ICA can reveal structure that is difficult to characterize through vocabulary-space projections alone.
Relative spatial context and texture are further candidates for this analysis, although we did not evaluate them here.

\subsection{Implications of Multimodal Training}
\label{sec:discussion_multimodal_training}

When a pretrained vision encoder is connected to a pretrained LM, a central question is how to map visual features into a space the LM can use. LLaVA's feature-alignment stage addresses this question by training the projector.
When the LM itself is trained on large-scale multimodal data, as in Qwen3 and Gemma3, it may also learn computations specific to the visual tokens it receives.

The layerwise changes in object- and position-selective axes are consistent with continued processing of visual tokens inside the LM. Object categories, spatial layout, and, in the qualitative examples, color are all represented within the same residual stream.
One possibility is that multimodal training shapes how these properties are organized across LM layers.

The model differences in \cref{sec:axes_dynamics} suggest possible contributions from architecture as well as training. For example, Qwen's dynamic-resolution tokenization and positional encoding may help preserve spatial structure, while differences in how models attend to visual tokens may affect the prominence of object-selective directions.
These features vary together across the models, so their individual contributions remain open.

\subsection{Limitations}
\label{sec:limitations}

This work analyzes representational structure rather than causal or behavioral mechanisms.
An ICA direction associated with an object category, position, or color indicates that the corresponding information is linearly accessible along that direction, but does not establish that the model causally uses it for generation.
Connecting the discovered components to downstream capabilities would require interventions or task-level evaluations.

The analysis also depends on the data distribution, ICA configuration, and input setting.
Our quantitative interpretation relies primarily on COCO annotations, and ICA components are not unique or exhaustive.
The present experiments do not establish that individual components retain the same interpretation on unseen datasets or categories.
In addition, text preceding the image could affect the analyzed visual-token states, so robustness to such conditioning remains an open question.
The layerwise differences may reflect architecture or training. The compared models differ simultaneously in visual tokenization, positional encoding, attention structure, training data, and other design choices. Controlled comparisons would be required to isolate their effects.

\section{CONCLUSION}
\label{sec:conclusion}

We applied ICA to visual-token representations in the LM residual streams of three LVLMs and interpreted each axis using the image regions associated with its highest-scoring tokens.
The fractions of object- and position-selective axes varied across layers, with distinct profiles across models.
Qualitative examples also showed axes whose highest-scoring regions shared a color.
By linking axes directly to visual properties such as token position, ICA complements vocabulary-space projections.
Future work could test whether intervening on these axes changes model outputs and whether the layerwise profiles generalize to new datasets.

\section*{ACKNOWLEDGEMENTS}

This work was supported by JSPS KAKENHI Grant Number JP26K00002.

We used OpenAI's ChatGPT for English editing throughout the manuscript and Codex to help implement and refine the experimental code. We reviewed and verified the AI-assisted content and take full responsibility for the manuscript and research results. 

{
\small
\bibliographystyle{apalike}
\bibliography{reference}

@inproceedings{
liu2023visual,
title={Visual Instruction Tuning},
author={Haotian Liu and Chunyuan Li and Qingyang Wu and Yong Jae Lee},
booktitle={Thirty-seventh Conference on Neural Information Processing Systems},
year={2023},
url={https://openreview.net/forum?id=w0H2xGHlkw},
  note={\url{https://openreview.net/forum?id=w0H2xGHlkw}}
}

@inproceedings{liu2024improved,
  author = {Haotian Liu and Chunyuan Li and Yuheng Li and Yong Jae Lee},
  title = {Improved Baselines with Visual Instruction Tuning},
  booktitle = {Proceedings of the IEEE/CVF Conference on Computer Vision and Pattern Recognition (CVPR)},
  month = jun,
  year = {2024},
  pages = {26296--26306},
  doi = {10.1109/CVPR52733.2024.02484},
  url = {https://openaccess.thecvf.com/content/CVPR2024/html/Liu_Improved_Baselines_with_Visual_Instruction_Tuning_CVPR_2024_paper.html},
  note = {doi: \href{https://doi.org/10.1109/CVPR52733.2024.02484}{\nolinkurl{10.1109/CVPR52733.2024.02484}}}
}

@misc{bai2025qwen3vltechnicalreport,
      title={{Qwen3-VL} Technical Report}, 
      author={Shuai Bai and Yuxuan Cai and Ruizhe Chen and Keqin Chen and Xionghui Chen and Zesen Cheng and others},
      year={2025},
      eprint={2511.21631},
      archivePrefix={arXiv},
      primaryClass={cs.CV},
      url={https://arxiv.org/abs/2511.21631},
  doi={10.48550/arXiv.2511.21631},
  note={doi: \href{https://doi.org/10.48550/arXiv.2511.21631}{\nolinkurl{10.48550/arXiv.2511.21631}}}
}

@misc{gemmateam2025gemma3technicalreport,
      collaboration={{Gemma Team}},
      title={{Gemma 3} Technical Report}, 
      author={{Gemma Team}},
      year={2025},
      eprint={2503.19786},
      archivePrefix={arXiv},
      primaryClass={cs.CL},
      url={https://arxiv.org/abs/2503.19786},
  doi={10.48550/arXiv.2503.19786},
  note={doi: \href{https://doi.org/10.48550/arXiv.2503.19786}{\nolinkurl{10.48550/arXiv.2503.19786}}}
}

@misc{logitlens,
  title = {Interpreting {{GPT}}: The Logit Lens},
  shorttitle = {Interpreting {{GPT}}},
  author = {{nostalgebraist}},
  year = {2020},
  month = Aug,
  url = {https://www.lesswrong.com/posts/AcKRB8wDpdaN6v6ru/interpreting-gpt-the-logit-lens},
  note = {\url{https://www.lesswrong.com/posts/AcKRB8wDpdaN6v6ru/interpreting-gpt-the-logit-lens}},
  urldate = {2025-09-18}
}

@inproceedings{
neo2025towards,
title={Towards Interpreting Visual Information Processing in Vision-Language Models},
author={Clement Neo and Luke Ong and Philip Torr and Mor Geva and David Krueger and Fazl Barez},
booktitle={The Thirteenth International Conference on Learning Representations},
year={2025},
url={https://openreview.net/forum?id=chanJGoa7f},
  note={\url{https://openreview.net/forum?id=chanJGoa7f}}
}

@misc{venhoff2025visualrepresentationsmaplanguage,
      title={How Visual Representations Map to Language Feature Space in Multimodal {LLMs}}, 
      author={Constantin Venhoff and Ashkan Khakzar and Sonia Joseph and Philip Torr and Neel Nanda},
      year={2025},
      eprint={2506.11976},
      archivePrefix={arXiv},
      primaryClass={cs.CV},
      url={https://arxiv.org/abs/2506.11976},
  doi={10.48550/arXiv.2506.11976},
  note={doi: \href{https://doi.org/10.48550/arXiv.2506.11976}{\nolinkurl{10.48550/arXiv.2506.11976}}}
}

@INPROCEEDINGS{zhang2025crossmodal,
  author={Zhang, Zhi and Yadav, Srishti and Han, Fengze and Shutova, Ekaterina},
  booktitle={2025 IEEE/CVF Conference on Computer Vision and Pattern Recognition (CVPR)}, 
  title={Cross-modal Information Flow in Multimodal Large Language Models}, 
  year={2025},
  volume={},
  number={},
  pages={19781-19791},
  doi={10.1109/CVPR52734.2025.01842},
  note={doi: \href{https://doi.org/10.1109/CVPR52734.2025.01842}{\nolinkurl{10.1109/CVPR52734.2025.01842}}}
}

@inproceedings{
serra2025the,
title={The Narrow Gate: Localized Image-Text Communication in Native Multimodal Models},
author={Alessandro Pietro Serra and Francesco Ortu and Emanuele Panizon and Lucrezia Valeriani and Lorenzo Basile and Alessio Ansuini and others},
booktitle={The Thirty-ninth Annual Conference on Neural Information Processing Systems},
year={2025},
url={https://openreview.net/forum?id=n0I0IvdfB3},
note={\url{https://openreview.net/forum?id=n0I0IvdfB3}}
}

@article{HYVARINEN2000411,
title = {Independent component analysis: algorithms and applications},
journal = {Neural Networks},
volume = {13},
number = {4--5},
pages = {411-430},
year = {2000},
issn = {0893-6080},
doi = {https://doi.org/10.1016/S0893-6080(00)00026-5},
url = {https://www.sciencedirect.com/science/article/pii/S0893608000000265},
author = {A. Hyvärinen and E. Oja},
  note={doi: \href{https://doi.org/10.1016/S0893-6080(00)00026-5}{\nolinkurl{10.1016/S0893-6080(00)00026-5}}}
}

@inproceedings{yamagiwa-etal-2023-discovering,
    title = "Discovering Universal Geometry in Embeddings with {ICA}",
    author = "Yamagiwa, Hiroaki  and
      Oyama, Momose  and
      Shimodaira, Hidetoshi",
    booktitle = "Proceedings of the 2023 Conference on Empirical Methods in Natural Language Processing",
    month = dec,
    year = "2023",
    publisher = "Association for Computational Linguistics",
    url = "https://aclanthology.org/2023.emnlp-main.283/",
    doi = "10.18653/v1/2023.emnlp-main.283",
    pages = "4647--4675",
  note={doi: \href{https://doi.org/10.18653/v1/2023.emnlp-main.283}{\nolinkurl{10.18653/v1/2023.emnlp-main.283}}}
}

@inproceedings{vaswani2017attention,
author = {Vaswani, Ashish and Shazeer, Noam and Parmar, Niki and Uszkoreit, Jakob and Jones, Llion and Gomez, Aidan N. and others},
title = {Attention is all you need},
year = {2017},
isbn = {9781510860964},
publisher = {Curran Associates Inc.},
booktitle = {Proceedings of the 31st International Conference on Neural Information Processing Systems},
pages = {6000--6010},
numpages = {11},
location = {Long Beach, California, USA},
series = {NIPS'17},
url = {https://proceedings.neurips.cc/paper/7181-attention-is-all-you-need},
note = {\url{https://proceedings.neurips.cc/paper/7181-attention-is-all-you-need}}
}

@inproceedings{xiong2020onlayer,
author = {Xiong, Ruibin and Yang, Yunchang and He, Di and Zheng, Kai and Zheng, Shuxin and Xing, Chen and others},
title = {On layer normalization in the transformer architecture},
year = {2020},
publisher = {PMLR},
booktitle = {Proceedings of the 37th International Conference on Machine Learning},
volume = {119},
pages = {10524--10533},
series = {Proceedings of Machine Learning Research},
url = {https://proceedings.mlr.press/v119/xiong20b.html},
note = {\url{https://proceedings.mlr.press/v119/xiong20b.html}}
}

@misc{mathematicalframework,
    title = {{A Mathematical Framework} for {Transformer Circuits}},
    author={Elhage, Nelson and Nanda, Neel and Olsson, Catherine and Henighan, Tom and Joseph, Nicholas and Mann, Ben and others},
    journal = {Transformer Circuits},
    year = {2021},
    urldate = {2025-09-20},
    url = {https://transformer-circuits.pub/2021/framework/index.html},
  note={\url{https://transformer-circuits.pub/2021/framework/index.html}}
}

@inproceedings{
huben2024sparse,
title={Sparse Autoencoders Find Highly Interpretable Features in Language Models},
author={Robert Huben and Hoagy Cunningham and Logan Riggs Smith and Aidan Ewart and Lee Sharkey},
booktitle={The Twelfth International Conference on Learning Representations},
year={2024},
url={https://openreview.net/forum?id=F76bwRSLeK},
  note={\url{https://openreview.net/forum?id=F76bwRSLeK}}
}

@inproceedings{
dunefsky2024transcoders,
title={Transcoders find interpretable {LLM} feature circuits},
author={Jacob Dunefsky and Philippe Chlenski and Neel Nanda},
booktitle={The Thirty-eighth Annual Conference on Neural Information Processing Systems},
year={2024},
url={https://openreview.net/forum?id=J6zHcScAo0},
  note={\url{https://openreview.net/forum?id=J6zHcScAo0}}
}

@inproceedings{oyama-etal-2024-understanding,
    title = "Understanding Higher-Order Correlations Among Semantic Components in Embeddings",
    author = "Oyama, Momose  and
      Yamagiwa, Hiroaki  and
      Shimodaira, Hidetoshi",
    booktitle = "Proceedings of the 2024 Conference on Empirical Methods in Natural Language Processing",
    month = nov,
    year = "2024",
    publisher = "Association for Computational Linguistics",
    url = "https://aclanthology.org/2024.emnlp-main.169/",
    doi = "10.18653/v1/2024.emnlp-main.169",
    pages = "2883--2899",
  note={doi: \href{https://doi.org/10.18653/v1/2024.emnlp-main.169}{\nolinkurl{10.18653/v1/2024.emnlp-main.169}}}
}

@inproceedings{wendler-etal-2024-llamas,
    title = "Do Llamas Work in {E}nglish? On the Latent Language of Multilingual Transformers",
    author = "Wendler, Chris  and
      Veselovsky, Veniamin  and
      Monea, Giovanni  and
      West, Robert",
    booktitle = "Proceedings of the 62nd Annual Meeting of the Association for Computational Linguistics (Volume 1: Long Papers)",
    month = aug,
    year = "2024",
    publisher = "Association for Computational Linguistics",
    url = "https://aclanthology.org/2024.acl-long.820/",
    doi = "10.18653/v1/2024.acl-long.820",
    pages = "15366--15394",
  note={doi: \href{https://doi.org/10.18653/v1/2024.acl-long.820}{\nolinkurl{10.18653/v1/2024.acl-long.820}}}
}

@inproceedings{
liu2025mechanistic,
title={Mechanistic Interpretability Meets Vision Language Models: Insights and Limitations},
author={Yiming Liu and Yuhui Zhang and Serena Yeung-Levy},
booktitle={The Fourth Blogpost Track at ICLR 2025},
year={2025},
url={https://openreview.net/forum?id=pZqvfsUpeh},
  note={\url{https://openreview.net/forum?id=pZqvfsUpeh}}
}

@InProceedings{mscoco,
author="Lin, Tsung-Yi and Maire, Michael and Belongie, Serge and Hays, James and Perona, Pietro and Ramanan, Deva and others",
title="Microsoft COCO: Common Objects in Context",
booktitle="Computer Vision -- ECCV 2014",
year="2014",
publisher="Springer International Publishing",
pages="740--755",
isbn="978-3-319-10602-1",
doi="10.1007/978-3-319-10602-1_48",
note={doi: \href{https://doi.org/10.1007/978-3-319-10602-1_48}{\nolinkurl{10.1007/978-3-319-10602-1_48}}}
}

@inproceedings{wolf-etal-2020-transformers,
    title = "Transformers: State-of-the-Art Natural Language Processing",
    author = "Wolf, Thomas and Debut, Lysandre and Sanh, Victor and Chaumond, Julien and Delangue, Clement and Moi, Anthony and others",
    booktitle = "Proceedings of the 2020 Conference on Empirical Methods in Natural Language Processing: System Demonstrations",
    month = oct,
    year = "2020",
    publisher = "Association for Computational Linguistics",
    url = "https://aclanthology.org/2020.emnlp-demos.6/",
    doi = "10.18653/v1/2020.emnlp-demos.6",
    pages = "38--45",
  note={doi: \href{https://doi.org/10.18653/v1/2020.emnlp-demos.6}{\nolinkurl{10.18653/v1/2020.emnlp-demos.6}}}
}

@inproceedings{kaduri-et-al-2025-deep-dive,
  author={Kaduri, Omri and Bagon, Shai and Dekel, Tali},
  booktitle={2025 IEEE/CVF Conference on Computer Vision and Pattern Recognition (CVPR)}, 
  title={What's in the Image? A Deep-Dive into the Vision of Vision Language Models}, 
  year={2025},
  volume={},
  number={},
  pages={14549-14558},
  doi={10.1109/CVPR52734.2025.01356},
  note={doi: \href{https://doi.org/10.1109/CVPR52734.2025.01356}{\nolinkurl{10.1109/CVPR52734.2025.01356}}}
}

@inproceedings{phukan-etal-2025-beyond,
    title = "Beyond Logit Lens: Contextual Embeddings for Robust Hallucination Detection {\&} Grounding in {VLM}s",
    author = "Phukan, Anirudh  and
      Divyansh  and
      Morj, Harshit Kumar  and
      Vaishnavi  and
      Saxena, Apoorv  and
      Goswami, Koustava",
    booktitle = "Proceedings of the 2025 Conference of the Nations of the Americas Chapter of the Association for Computational Linguistics: Human Language Technologies (Volume 1: Long Papers)",
    month = apr,
    year = "2025",
    publisher = "Association for Computational Linguistics",
    url = "https://aclanthology.org/2025.naacl-long.488/",
    doi = "10.18653/v1/2025.naacl-long.488",
    pages = "9661--9675",
    ISBN = "979-8-89176-189-6",
  note={doi: \href{https://doi.org/10.18653/v1/2025.naacl-long.488}{\nolinkurl{10.18653/v1/2025.naacl-long.488}}}
}
}

\end{document}